\documentclass{article}

\usepackage[preprint]{neurips_2025}

\usepackage[utf8]{inputenc} 
\usepackage[T1]{fontenc}    
\usepackage[colorlinks,linkcolor=red,anchorcolor=blue,citecolor=green]{hyperref}       
\usepackage{url}            
\usepackage{booktabs}       
\usepackage{amsfonts}       
\usepackage{nicefrac}       
\usepackage{microtype}      
\usepackage{enumitem}
\usepackage{pifont}
\usepackage{graphicx}
\usepackage{booktabs}
\usepackage{multirow}
\usepackage{graphicx}
\usepackage[table,xcdraw]{xcolor}
\usepackage{amsmath}
\numberwithin{equation}{section}
\usepackage{algorithmic}
\usepackage{algorithm}
\usepackage{wrapfig}
\usepackage[most]{tcolorbox}
\usepackage{lipsum} 
\usepackage{titlesec} %
\usepackage{subcaption}

\newcommand{\Tref}[1]{Tab.~\ref{#1}}

\newcommand{\Fref}[1]{Fig.~\ref{#1}}
\newcommand{\Sref}[1]{Sec.~\ref{#1}}

\newcommand{\ie}{\textit{i}.\textit{e}.}
\newcommand{\eg}{\textit{e}.\textit{g}.}

\newcommand{\vs}{\textit{vs}. }

\newcommand{\toolns}{\textsc{Bench2ADVLM}}
\newcommand{\tool}{\toolns\space}
\newcommand{\advlmns}{ADVLMs}
\newcommand{\advlm}{\advlmns\space}

\newcommand*{\affaddr}[1]{#1} 
\newcommand*{\affmark}[1][*]{\textsuperscript{#1}}

\title{\toolns: A Closed-Loop Benchmark for Vision-language Models in Autonomous Driving}

\author{
    Tianyuan Zhang\affmark[1],
    Ting Jin\affmark[1], 
    Lu Wang\affmark[1], 
    Jiangfan Liu\affmark[1], 
    Siyuan Liang\affmark[2], \\
    \textbf{Mingchuan Zhang\affmark[3],} 
    \textbf{Aishan Liu\affmark[1]} \thanks{Corresponding Author}
    \textbf{, Xianglong Liu\affmark[1]}
    \\
    \affaddr{\affmark[1]{Beihang University}}
    \affaddr{\affmark[2]{Nanyang Technological University}}
    \\
    \affaddr{\affmark[3]{Henan University of Science and Technology}} 
}

\begin{document}

\maketitle

\begin{abstract}
Vision-Language Models (VLMs) have recently emerged as a promising paradigm in autonomous driving (AD). However, current performance evaluation protocols for VLM-based AD systems (\advlmns) are predominantly confined to open-loop settings with static inputs, neglecting the more realistic and informative closed-loop setting that captures interactive behavior, feedback resilience, and real-world safety. To address this, we introduce \toolns, a unified hierarchical closed-loop evaluation framework for real-time, interactive assessment of \advlm across both simulation and physical platforms. Inspired by dual-process theories of cognition, we first adapt diverse \advlm to simulation environments via a dual-system adaptation architecture. In this design, heterogeneous high-level driving commands generated by target \advlm (fast system) are interpreted by a general-purpose VLM (slow system) into standardized mid-level control actions suitable for execution in simulation. To bridge the gap between simulation and reality, we design a physical control abstraction layer that translates these mid-level actions into low-level actuation signals, enabling, for the first time, closed-loop testing of \advlm on physical vehicles. To enable more comprehensive evaluation, \tool introduces a self-reflective scenario generation module that automatically explores model behavior and uncovers potential failure modes for safety-critical scenario generation. Overall, \tool establishes a hierarchical evaluation pipeline that seamlessly integrates high-level abstract reasoning, mid-level simulation actions, and low-level real-world execution. Experiments on diverse scenarios across multiple state-of-the-art \advlm and physical platforms validate the diagnostic strength of our framework, revealing that existing \advlm still exhibit limited performance under closed-loop conditions. To our knowledge, this is the first work to establish the closed-loop evaluation benchmark for \advlmns, offering a principled path toward scalable, reliable deployment of \advlmns. 
\end{abstract}

\section{Introduction}
\label{sec:introduction}

With the rapid advancement of deep learning, autonomous driving (AD) technology has progressed from modular pipelines to end-to-end systems \cite{hu2023planning,jiang2023vad,chen2024end,wu2022trajectory,zhang2021end}, and more recently, to Vision-Language Models (VLMs) \cite{achiam2023gpt,alayrac2022flamingo,liu2024llava,team2023gemini}. Owing to their strong generalization capabilities and enhanced interpretability, VLMs have emerged as a promising paradigm in contemporary AD research \cite{wu2025language,xu2024drivegpt4,nie2024reason2drive,dolphins,drivelm,fu2025orion, zhang2024visual, wang2025black}.

Though promising, the evaluations of VLM-based AD systems (\advlmns) remain critically limited as they primarily focus on the open-loop setting, where the model outputs are not fed back into the AD system in the simulation environment, resulting in static inputs that do not reflect the consequences of the model's actions. However, the more reliable and difficult closed-loop evaluation \cite{jia2024bench2drive} has not been explored, which enables dynamic and real-time feedback with the environment by continuously incorporating outputs into future inputs. Compared to open-loop evaluations, closed-loop testing can effectively mitigate issues such as distribution shift \cite{ross2011reduction} and cascading errors \cite{de2019causal,jia2023driveadapter} caused by the accumulation of incorrect predictions, which may help to reveal more flaws of AD systems.


A fundamental \emph{challenge} in the closed-loop evaluation of current \advlm lies in their inability to interact with the physical environment directly. In contrast to traditional end-to-end driving models, \advlm operate at a high level of abstraction, generating semantic driving commands rather than executable control signals. This architectural disconnect prevents direct interfacing with actuation systems and necessitates an additional interpretation layer to translate abstract decisions into control signals. As a result, the closed-loop feedback cycle is disrupted, limiting the fidelity and effectiveness of simulation-based and real-world evaluations. In addition, existing research predominantly focuses on virtual environments, with limited efforts toward physical-world validation. Physical deployment remains largely unexplored, and even small-scale real-vehicle testing is rarely conducted, despite its necessity for assessing real-world reliability and safety~\cite{liu2025natural,liang2024object,liang2020efficient,liang2022parallel,liang2022large,liu2023x}.

To address the issue, we propose \toolns, a unified closed-loop framework for real-time and interactive assessment of \advlm in both virtual and physical environments. Inspired by the dual-system theory of fast and slow cognition in psychology~\cite{kahneman2011thinking}, \tool adopts a dual-system adaptation architecture that simulates the roles of a ``fast system'' and a ``slow system''. The fast system (\ie, the target \advlm under evaluation) generates high-level driving commands, while the slow system (\ie, the general-purpose VLMs) acts as a semantic executor that translates these commands into mid-level control actions executable within the simulation environment.
Furthermore, we establish a physical control abstraction layer that maps mid-level control actions to low-level real-world actuation signals, enabling, for the first time, closed-loop testing of \advlm on physical vehicles. In summary, \tool offers the first unified testbed that hierarchically connects abstract language-based reasoning with both simulation control and physical actuation, closing the loop across all layers of AD evaluation. 
Additionally, to support customized evaluation of target models, \tool incorporates a self-reflective scenario generation module built upon the routes provided in the Bench2Drive \cite{jia2024bench2drive}, autonomously generating 220 threat scenarios, enabling the \advlm to be exposed to a wider spectrum of adversarial, long-tail, and safety-critical conditions~\cite{zhang2024lanevil,lu2025adversarial,kong2024environmental,zhang2024visual}. At last, we discussed the potential directions for designing advanced \advlm based on our insights. We will continuously develop this ecosystem for the community. The main \textbf{contributions} are:

\begin{itemize}
    \item \textbf{Dual-system Adaptation Architecture.} We propose a dual-system adaptation architecture inspired by fast and slow cognition, where the target \advlm generates high-level driving commands, the general-purpose VLMs translate them into mid-level control actions, enabling closed-loop testing of \advlm in simulators.

    \item \textbf{Physical Control Abstraction Layer.} We build a physical control abstraction layer to map the mid-level control actions to real-world actuation signals, enabling closed-loop testing of \advlm on physical vehicles.
    
   \item \textbf{Self-Reflective Scenario Generation.} We introduce a self-reflective scenario generation module that, based on 220 standard routes, autonomously creates 220 safety-critical scenarios, enabling more diverse and targeted evaluation.
    
\end{itemize}

\section{Related Work}
\label{sec:related_work}




\textbf{VLM-based Autonomous Driving.} Recent works explore various \advlm \cite{wu2025language, ding2023hilm, najibi2023unsupervised, xu2024drivegpt4, nie2024reason2drive}. EM-VLM4AD~\cite{gopalkrishnanmulti} improves efficiency through a lightweight architecture without compromising VQA performance. Dolphins~\cite{dolphins} further enables human-like interaction using a Grounded Chain of Thought (GCoT) and in-context instruction tuning.
GPT-Driver~\cite{mao2023gpt} reformulates motion planning as a language modeling task. LLM-driver~\cite{chen2024driving} introduces a multimodal architecture that merges vectorized object-level representations with LLMs. MTD-GPT~\cite{liu2023mtd} integrates reinforcement learning with GPT-based sequence modeling to support multi-task decision-making.
Further work, such as LanguageMPC~\cite{sha2310languagempc}, DriveVLM~\cite{tiandrivevlm}, and DriveMLM~\cite{wang2023drivemlm} explores VLMs as decision-makers and planners. Senna~\cite{jiang2024senna} proposes different types of QAs to improve the planning performance, and DriveLM~\cite{drivelm} unifies both aspects within a Graph Visual Question Answering (GVQA) framework, enabling the model to output textual answers to driving-related queries alongside future trajectory waypoints. OmniDrive \cite{wang2025omnidrive} bridges 3D driving perception and language-based reasoning through counterfactual supervision. ORION~\cite{fu2025orion} explores vision-language instructed action generation in closed-loop settings; however, the underlying decision-making process is still based on a generic generative model rather than the \advlmns.

Despite progress, most \advlm are only evaluated in open-loop settings. To capture decision-environment interaction, we focus on closed-loop evaluation in planning scenarios.


\textbf{Benchmarks of Autonomous Driving.}
\ding{182} \emph{Open-Loop Benchmarks for ADVLMs}: Most existing \advlmns' benchmarks focus on open-loop evaluation, emphasizing multimodal reasoning and task-specific performance, particularly in VQA. LingoQA~\cite{marcu2024lingoqa} introduces a large-scale dataset along with Lingo-Judge, a learned metric that surpasses traditional text-based scores such as BLEU~\cite{papineni2002bleu}, METEOR~\cite{banerjee2005meteor}, CIDEr~\cite{vedantam2015cider}, and GPT-4~\cite{achiam2023gpt}. Reason2Drive~\cite{nie2024reason2drive} provides over 600K video-text pairs from Waymo~\cite{sun2020scalability}, nuScenes~\cite{caesar2020nuscenes}, and ONCE~\cite{mao1one}, aiming to evaluate interpretable, chain-based reasoning via ADRScore. NuScenes-QA~\cite{qian2024nuscenes} adds 34K scenes with 460K QA pairs. AutoTrust~\cite{xing2024autotrust} builds upon a diverse data collection to assess the \advlm across various facets. Meanwhile, DriveBench~\cite{xie2025vlms} establishes a comprehensive benchmark to evaluate the reliability under diverse settings and tasks with refined evaluation metrics. 
Beyond VQA, benchmarks like DriveLM~\cite{drivelm} and Dolphins~\cite{dolphins} extend to planning tasks but remain open-loop, limiting real-world relevance in dynamic settings. \ding{183} \emph{Closed-Loop Benchmarks for end-to-end AD}: In closed-loop evaluation, existing benchmarks are primarily designed for traditional end-to-end AD models. CARLA~\cite{dosovitskiy2017carla} is the most widely used open-source simulator and has enabled the development of several representative closed-loop benchmarks. CARLA Leaderboard v1~\cite{CarlaLeaderboard}, introduced as an algorithm competition, evaluates basic navigation and simple interactive behaviors. Its successor, CARLA Leaderboard v2, expands the environment set with more complex scenarios and additional towns. Building on CARLA, Bench2Drive~\cite{jia2024bench2drive} proposes a more granular and comprehensive closed-loop benchmark tailored to end-to-end AD models. 

However, existing benchmarks assume direct low-level control signals and are incompatible with \advlmns. This gap underscores the need for a dedicated closed-loop \advlm benchmark.

\section{\tool Design}
\label{sec:benchmark}

\begin{figure}[!t]
    \centering
    \includegraphics[width=0.95\linewidth]{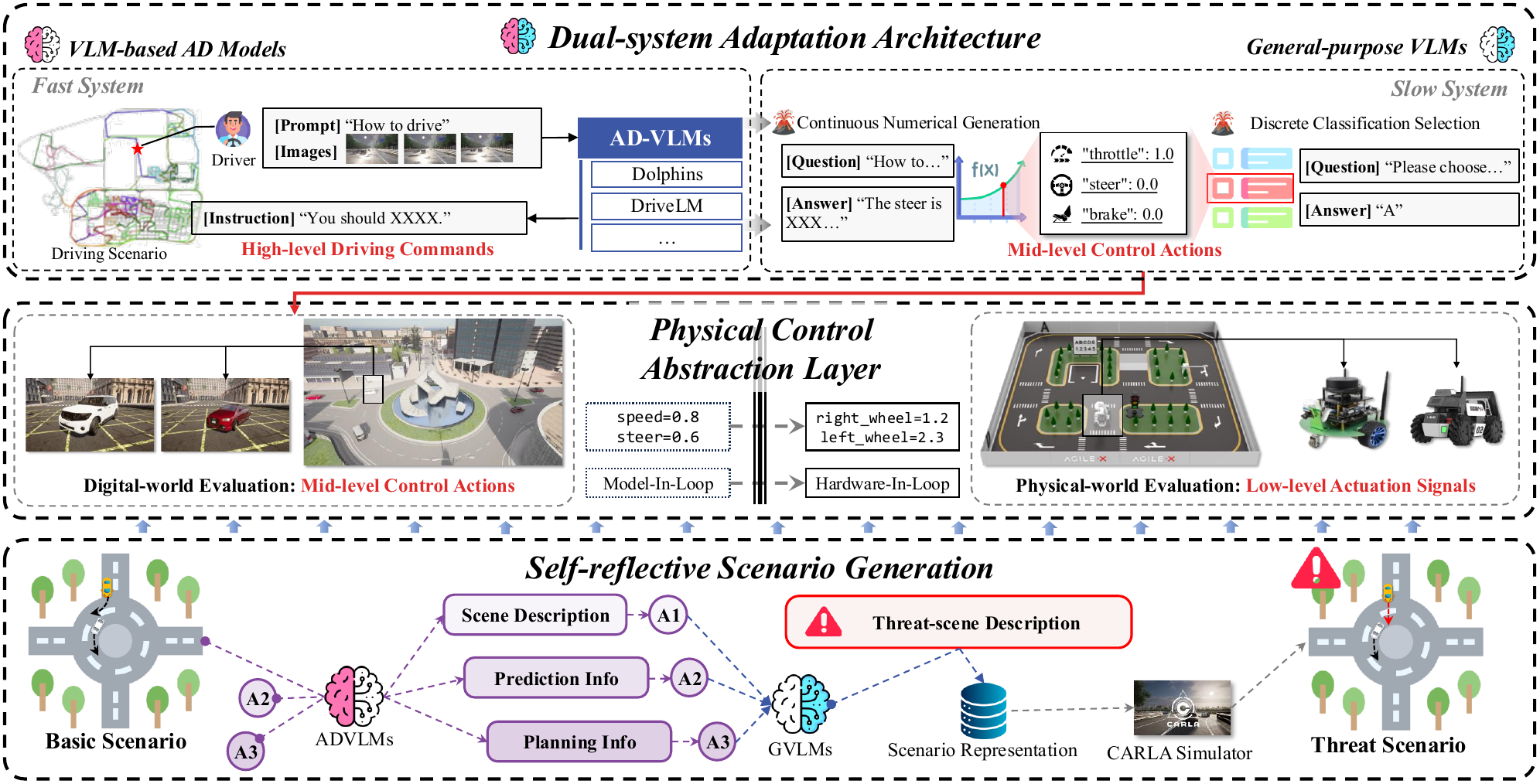}
    \caption{Overview of the \tool benchmark. The framework includes a dual-system adaptation architecture for translating high-level driving commands into mid-level control actions, a physical control abstraction layer for mapping mid-level control actions to low-level actuation signals, and a self-reflective scenario generation module for probing potential failure modes.}
    \label{fig:framework}
    \vspace{-0.12in}
\end{figure}

\tool is a unified closed-loop evaluation framework for real-time and interactive assessment of \advlmns. The framework of \tool is shown in \Fref{fig:framework}.

\subsection{Dual-system Adaptation Architecture}

Inspired by the dual-system theory of cognition in psychology~\cite{kahneman2011thinking}, we design a dual-system adaptation architecture that emulates the interaction between a fast and a slow cognitive process. In our framework, the \textit{fast system} corresponds to the target \advlmns, which is responsible for generating high-level, goal-directed driving commands based on visual-linguistic input. The \textit{slow system}, implemented using general-purpose VLMs (GVLMs), acts as a semantic executor that translates these high-level driving commands into executable mid-level control actions. 

We first introduce the \textit{fast system}, which corresponds to the target \advlmns. The outputs of different \advlm exhibit significant heterogeneity, including structured commands, trajectory predictions, and free-form textual reasoning. This diversity makes applying a unified pattern recognition method for interpretation challenging. To address this, we abstract the outputs and leverage the generalization capability of large models for downstream processing.
Given the input image sequence $\{\mathbf{I}_{t-k}, \dots, \mathbf{I}_t\}$ and task-specific prompts set $\{\mathbf{Q}_t^\tau\}_{\tau \in \mathcal{T}}$, the \advlm generate the task-conditioned textual outputs:

\begin{equation}
\mathbb{T}_t = \left\{ \mathbf{t}_t^{\tau} = \mathcal{F}_{\text{fast}}(\{\mathbf{I}_{t-k}, \dots, \mathbf{I}_t\},\; \mathbf{Q}_t^\tau) \;\middle|\; \tau \in \mathcal{T} \right\},
\end{equation}

\noindent
where $\{\mathbf{I}_{t-k}, \dots, \mathbf{I}_t\}$ is the input image sequence, $\mathbf{Q}_t^\tau$ is the prompt for task $\tau$, and $\mathcal{T}$ is the set of predefined tasks (\eg, action prediction, trajectory forecasting, semantic reasoning). $\mathcal{F}_{\text{fast}}$ is the shared reasoning function, and $\mathbb{T}_t$ collects the task-specific textual outputs at time $t$.

The \textit{slow system} serves as a semantic interpreter that transforms heterogeneous high-level driving commands into executable mid-level control actions. Given the textual outputs $\mathbb{T}_t$ generated by the fast system, the slow system bridges the gap between semantic intentions and physical actions. Specifically, for each action-related task $\tau$, the corresponding textual description $\mathbf{t}_t^\tau$ is translated into control commands compatible with the CARLA simulator interface. In AD control, key actuation signals include steering, throttle, and brake values, corresponding to lateral and longitudinal vehicle maneuvers. To transform heterogeneous high-level driving commands into executable mid-level control actions, we design a semantic-to-control translation module based on GVLMs, tailored to the CARLA control protocol. Given a task-specific textual output $\mathbf{t}_t^\tau$ and the current observation image sequence $\{\mathbf{I}_{t-k}, \dots, \mathbf{I}_t\}$, the slow system predicts a control command $\mathbf{u}_t \in \mathbb{R}^3$ through:

\begin{equation}
\mathbf{u}_t = \mathcal{F}_{\text{slow}}(\mathbf{t}_t^\tau,\; \{\mathbf{I}_{t-k}, \dots, \mathbf{I}_t\},\; \mathbf{P},\; \mathbb{U}),
\end{equation}

\noindent
where $\mathbf{P}$ is the predefined control prompt template and $\mathbb{U}$ denotes the candidate control set. The control vector $\mathbf{u}_t = (s_t, \gamma_t, b_t)$ consists of three components: steering command $s_t$, throttle command $\gamma_t$, and brake command $b_t$. These signals correspond to the actuation primitives in the vehicle control interface of CARLA, with $s_t \in [-1, 1]$ controlling the steering angle, and $\gamma_t, b_t \in [0, 1]$ modulating the acceleration and braking intensities, respectively.

For VLMs, the raw image sequence $\{\mathbf{I}_{t-k}, \dots, \mathbf{I}_t\}$ is directly provided as visual input alongside the prompt. For pure LLMs, the image sequence is first encoded into a textual description, which is appended to the task-specific output $\mathbf{t}_t^\tau$ within the prompt template $\mathbf{P}$.

We instantiate $\mathcal{F}_{\text{slow}}$ under two operational modes: \ding{182} Continuous Numerical Generation (CNG) mode, $\mathbb{U}$ is left empty, and the VLM directly regresses continuous control values. Each output component is constrained within CARLA's specification ranges (\eg, steering angle in $[-1,1]$, throttle and brake in $[0,1]$). \ding{183} Discrete Classification Selection (DCS) mode, $\mathbb{U}$ is predefined as a discrete set of candidate control vectors derived from domain knowledge. The VLM selects the most semantically aligned candidate based on the high-level driving command. In both modes, the resulting control actions $\mathbf{u}_t$ are fed into CARLA’s vehicle interface, enabling real-time actuation based on high-level semantic decisions.
\textit{Details of prompt construction $\mathbf{P}$ and candidate sets $\mathbb{U}$ are provided in Appendix.}

\subsection{Physical Control Abstraction Layer}
\begin{figure}
    \centering
    \includegraphics[width=0.95\linewidth]{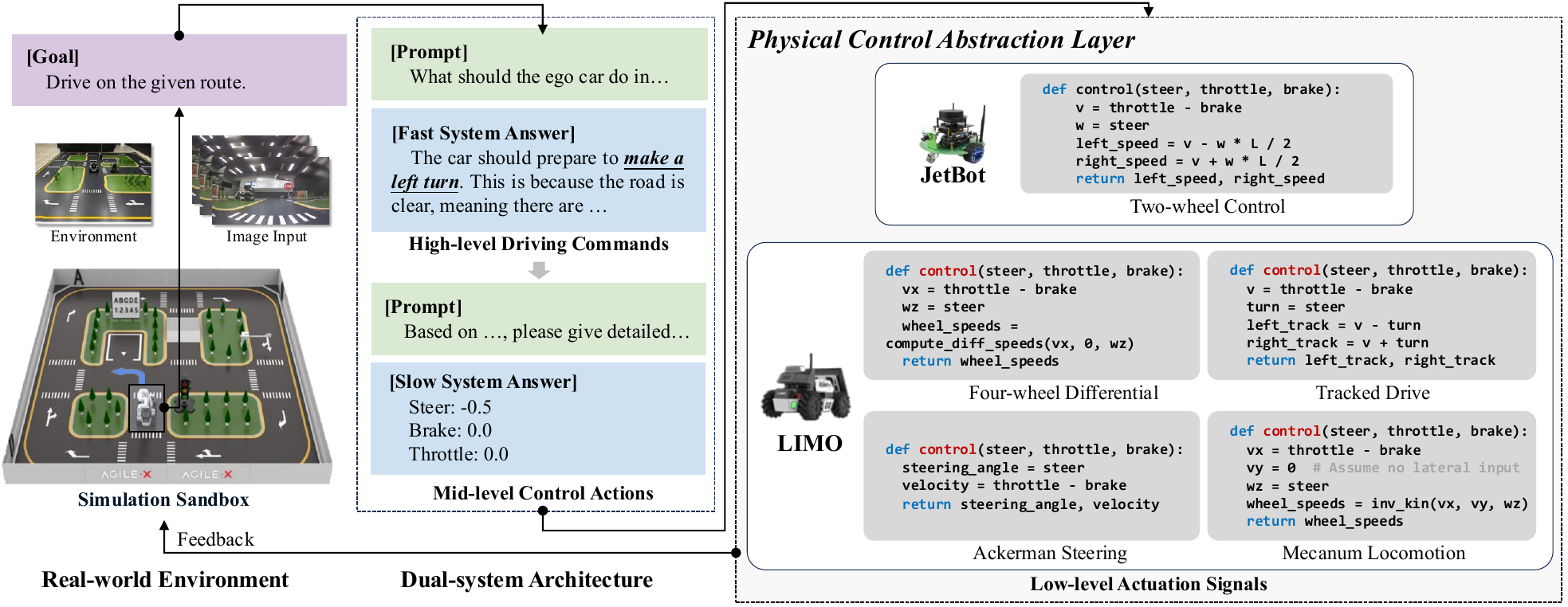}
    \caption{Illustration of the physical control abstraction layer. Physical-world evaluation is performed on the AGILE·X sandbox using Jetbot and LIMO. Vehicles collect real-time data and send it to the dual-system, which produces high-level commands and mid-level actions. These are then translated by the abstraction layer into platform-specific low-level actuation signals, closing the control loop.}
    \label{fig:virtual_layer}
    \vspace{-0.12in}
\end{figure}

To enable closed-loop evaluation in the physical world, we extend the testing paradigm from Model-in-the-Loop (MIL) simulation to Hardware-in-the-Loop (HIL) deployment. Specifically, we develop a physical control abstraction layer that maps mid-level control actions to low-level real-world actuation signals. At each control cycle, the system transmits the generated control signals to the physical vehicle, which executes the motion for a fixed duration of 0.5 seconds, captures the updated onboard observation, and feeds it back for the next inference step. This forms a real-time feedback loop across perception, decision-making, and actuation within the physical environment.

The physical control abstraction layer is validated in the AGILE·X simulation sandbox~\cite{limo} using two autonomous driving platforms: Jetbot~\cite{jetbot} and LIMO~\cite{limo}. Both platforms are equipped with onboard sensors, including cameras, LiDAR, and IMU, and are capable of standard motion control. Jetbot, featuring stronger onboard computational resources, is suited for AI-intensive workloads, while LIMO emphasizes actuation stability and supports multiple driving modes, including differential, Ackermann, tracked, and Mecanum configurations. Both platforms adopt ROS as the internal communication and control framework. To overcome the computational limitations of edge devices, we adopt a client-server architecture. The server hosts the vision-language planner and the decision-to-control parsing module, while the client, deployed onboard the vehicle, is responsible for real-time sensor acquisition and low-level actuation signals. Communication between the two sides is established via TCP sockets secured through SSH tunnels, ensuring low-latency and reliable feedback.

\begin{wrapfigure}{r}{0.5\linewidth}
    \centering
    \vspace{-0.12in}    
    \includegraphics[width=0.95\linewidth]{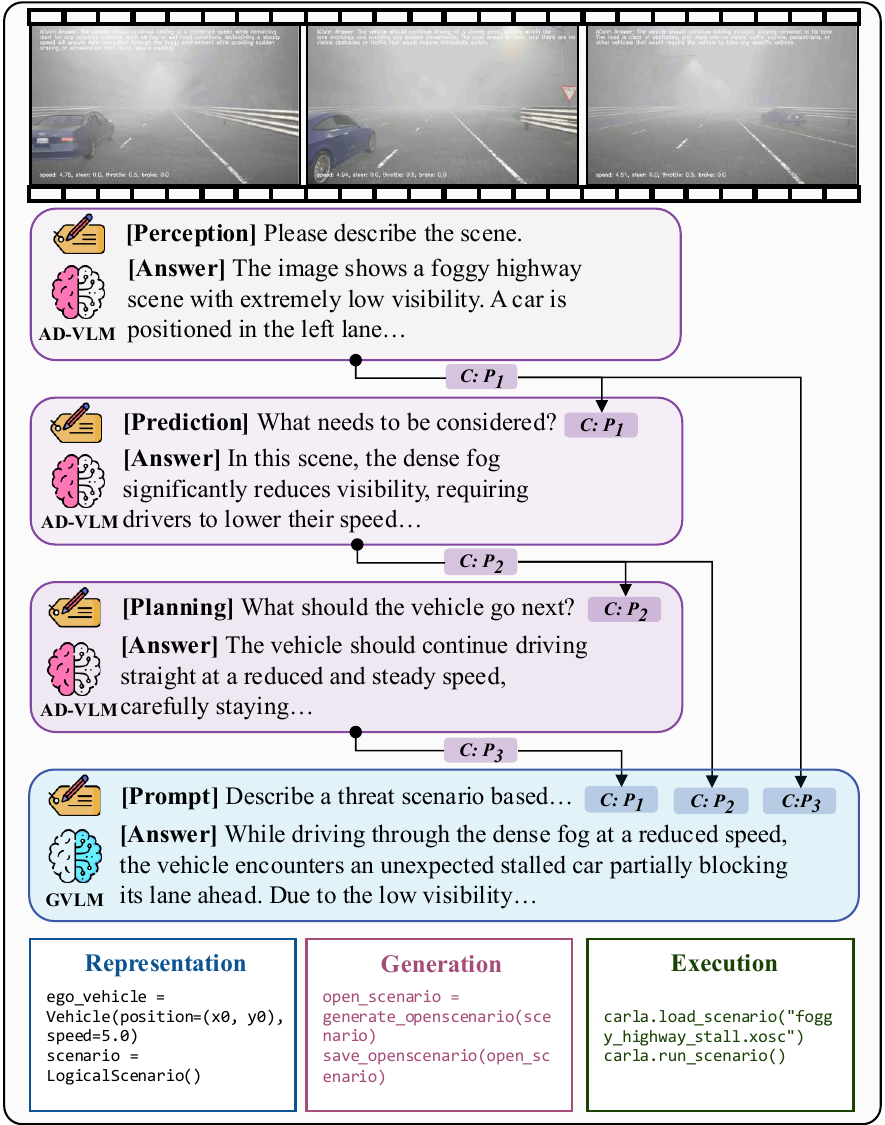}
    \caption{Self-reflective scenario generation. The ADVLM is prompted with P3 queries, and a GVLM fuses the answers into a description.}
    \label{fig:self-reflective}
    \vspace{-0.85in} 
\end{wrapfigure}

Notably, while our current deployment is demonstrated within the AGILE·X sandbox on Jetbot and LIMO, the modularity of the abstraction layer and the client-server separation design enable seamless extension to broader testing scenarios, supporting scalable transition from MIL to HIL settings.

\subsection{Self-reflective Scenario Generation}

Closed-loop evaluation requires an interactive environment and a diverse set of safety-critical scenarios that can challenge the target model. Existing benchmarks often rely on predefined patterns or rule-based perturbations to construct test cases, which may fail to capture the specific vulnerabilities of different \advlmns. This limits their ability to support targeted, diagnostic evaluation.

To address this, we introduce a self-reflective scenario generation mechanism that actively involves the target \advlm in the construction of threat-centric test scenarios, as shown in \Fref{fig:self-reflective}. Rather than passively applying static templates \cite{cai2025text2scenario}, we engage the model under test to elicit its own interpretation of the scene and the context it considers critical for decision-making. 

The generation process is organized around a structured three-stage reasoning framework inspired by the perception–prediction–planning (P3) paradigm widely used in AD~\cite{wang2025black}. Starting from basic scenarios, the system prompts the \advlm with queries at three cognitive stages: \ding{182} \textit{Perception}, describe what is happening now; \ding{183} \textit{Prediction}, identify influential factors for near-future evolution; and \ding{184} \textit{Planning}, determine the appropriate next action. Each response captures a distinct aspect of the model's internal reasoning. As \advlm focus on task-specific outputs, they struggle to unify cross-task signals. We address this by using a GVLM to fuse intermediate outputs into a coherent threat scene description. The resulting description is passed to a controllable scenario construction pipeline~\cite{cai2025text2scenario} and used as a prompt to instantiate simulation environments. Based on the 220 standard routes provided by the Bench2Drive \cite{jia2024bench2drive}, we generate a corresponding set of 220 threat-critical scenarios using our self-reflective generation method.
\section{Experiments}
\label{sec:experiments}

\subsection{Experimental Setup}
\label{sec:Experimental Setup}


\textbf{Models.} 
As the fast system, we evaluate four commonly-used \advlmns: Dolphins~\cite{dolphins}, DriveLM~\cite{drivelm}, EM-VLM4AD \cite{gopalkrishnanmulti} and OmniDrive \cite{wang2025omnidrive}. 
For the slow system, we compare a LLM and a VLM by using LLaMA-3-8B~\cite{touvron2023llama} and LLaVA-1.5-13B~\cite{liu2024llava}. \textbf{Parsing Mode.}  
Each model is tested under two parsing modes, Continuous Numerical Generation (CNG) and Discrete Classification Selection (DCS), as described in \Sref{sec:benchmark}. \textbf{Hardware Configuration.} A server equipped with a 128-core Intel Xeon 8358 CPU (2.60GHz), 1TB RAM, and 8×NVIDIA A800 80GB PCIe GPUs.

\textbf{Evaluation Metrics.} We construct a progressive evaluation framework by integrating metrics from the CARLA Leaderboard~\cite{CarlaLeaderboard} and Bench2Drive~\cite{jia2024bench2drive}, covering three dimensions: basic performance, measured by \textit{Success Rate}~$\textcolor{blue}\uparrow$ (percentage of routes completed without violations) and \textit{Driving Score}~$\textcolor{blue}\uparrow$ (completion score weighted by penalties); behavioral quality, assessed via \textit{Efficiency}~$\textcolor{blue}\uparrow$ (relative speed sampled every 5\% of the route) and \textit{Comfort}~$\textcolor{blue}\uparrow$ (proportion of smooth 20-frame segments with stable control signals); and specialized capability, evaluated by a \textit{Skill Score}~$\textcolor{blue}\uparrow$ averaged across five interactive skills: lane merging, overtaking, emergency braking, yielding, and traffic sign recognition. Each experiment is repeated 10 times, and the average results are reported.


\subsection{Main Results}


\label{sec:Main Results}

\begin{table}[t]
\centering
\caption{Main experimental results on \toolns. $^\dagger$ and $^\star$ indicate the use of Continuous Numerical Generation and Discrete Classification Selection parsing modes, respectively. Blue subscripts denote the standard deviation (\textcolor{blue}{±std}) over multiple runs.}
\label{tab:main}
\resizebox{\textwidth}{!}{%
\begin{tabular}{@{}c|c|cc|cc|c@{}}
\toprule[1.0pt]
\multirow{2}{*}{\textbf{Fast System}} & \multirow{2}{*}{\textbf{Slow System}} & \multicolumn{2}{c|}{\textbf{Basic Performance}} & \multicolumn{2}{c|}{\textbf{Behavior Quality}} & \textbf{Specialized Capability} \\ \cmidrule(l){3-7} 
 &  & \textit{Success Rate} $\textcolor{blue}\uparrow$ & \textit{Driving Score} $\textcolor{blue}\uparrow$ & \textit{Efficiency} $\textcolor{blue}\uparrow$ & \textit{Comfortness} $\textcolor{blue}\uparrow$ & \textit{Skill Score} $\textcolor{blue}\uparrow$ \\ \midrule
\multirow{4}{*}{DriveLM \cite{drivelm}} 
& LLaMA-3-8B$^\dagger$ & 9.09\textsubscript{$_\text{\textcolor{blue}{±0.42}}$} & 39.41\textsubscript{$_\text{\textcolor{blue}{±0.68}}$} & 127.55\textsubscript{$_\text{\textcolor{blue}{±2.13}}$} & 31.95\textsubscript{$_\text{\textcolor{blue}{±1.62}}$} & 17.55\textsubscript{$_\text{\textcolor{blue}{±0.38}}$} \\
& LLaMA-3-8B$^\star$ & 10.45\textsubscript{$_\text{\textcolor{blue}{±0.49}}$} & 38.79\textsubscript{$_\text{\textcolor{blue}{±0.75}}$} & 107.44\textsubscript{$_\text{\textcolor{blue}{±1.90}}$} & 65.20\textsubscript{$_\text{\textcolor{blue}{±2.10}}$} & 16.60\textsubscript{$_\text{\textcolor{blue}{±0.44}}$} \\ \cmidrule(l){2-7} 
& LLaVA-1.5-13B$^\dagger$ & 6.82\textsubscript{$_\text{\textcolor{blue}{±0.41}}$} & 37.29\textsubscript{$_\text{\textcolor{blue}{±0.63}}$} & 129.83\textsubscript{$_\text{\textcolor{blue}{±2.05}}$} & 28.95\textsubscript{$_\text{\textcolor{blue}{±1.94}}$} & 14.11\textsubscript{$_\text{\textcolor{blue}{±0.35}}$} \\
& LLaVA-1.5-13B$^\star$ & 5.00\textsubscript{$_\text{\textcolor{blue}{±0.38}}$} & 42.06\textsubscript{$_\text{\textcolor{blue}{±0.61}}$} & 85.74\textsubscript{$_\text{\textcolor{blue}{±1.76}}$} & 69.86\textsubscript{$_\text{\textcolor{blue}{±2.03}}$} & 13.10\textsubscript{$_\text{\textcolor{blue}{±0.41}}$} \\ \midrule
\multirow{4}{*}{Dolphins \cite{dolphins}} 
& LLaMA-3-8B$^\dagger$ & 9.09\textsubscript{$_\text{\textcolor{blue}{±0.46}}$} & 38.64\textsubscript{$_\text{\textcolor{blue}{±0.70}}$} & 125.72\textsubscript{$_\text{\textcolor{blue}{±1.85}}$} & 37.85\textsubscript{$_\text{\textcolor{blue}{±1.55}}$} & 15.73\textsubscript{$_\text{\textcolor{blue}{±0.32}}$} \\
& LLaMA-3-8B$^\star$ & 10.00\textsubscript{$_\text{\textcolor{blue}{±0.44}}$} & 34.32\textsubscript{$_\text{\textcolor{blue}{±0.66}}$} & 132.58\textsubscript{$_\text{\textcolor{blue}{±1.97}}$} & 43.54\textsubscript{$_\text{\textcolor{blue}{±1.73}}$} & 17.25\textsubscript{$_\text{\textcolor{blue}{±0.37}}$} \\ \cmidrule(l){2-7} 
& LLaVA-1.5-13B$^\dagger$ & 8.18\textsubscript{$_\text{\textcolor{blue}{±0.39}}$} & 33.62\textsubscript{$_\text{\textcolor{blue}{±0.62}}$} & 135.38\textsubscript{$_\text{\textcolor{blue}{±2.01}}$} & 50.68\textsubscript{$_\text{\textcolor{blue}{±1.61}}$} & 13.69\textsubscript{$_\text{\textcolor{blue}{±0.36}}$} \\
& LLaVA-1.5-13B$^\star$ & 2.73\textsubscript{$_\text{\textcolor{blue}{±0.36}}$} & 34.18\textsubscript{$_\text{\textcolor{blue}{±0.59}}$} & 72.46\textsubscript{$_\text{\textcolor{blue}{±1.83}}$} & 68.98\textsubscript{$_\text{\textcolor{blue}{±1.72}}$} & 6.26\textsubscript{$_\text{\textcolor{blue}{±0.39}}$} \\ \midrule
\multirow{4}{*}{EM-VLM4AD \cite{gopalkrishnanmulti}} 
& LLaMA-3-8B$^\dagger$ & 8.18\textsubscript{$_\text{\textcolor{blue}{±0.47}}$} & 36.29\textsubscript{$_\text{\textcolor{blue}{±0.65}}$} & 133.55\textsubscript{$_\text{\textcolor{blue}{±2.12}}$} & 28.75\textsubscript{$_\text{\textcolor{blue}{±1.85}}$} & 17.71\textsubscript{$_\text{\textcolor{blue}{±0.32}}$} \\
& LLaMA-3-8B$^\star$ & 10.00\textsubscript{$_\text{\textcolor{blue}{±0.43}}$} & 36.63\textsubscript{$_\text{\textcolor{blue}{±0.72}}$} & 123.53\textsubscript{$_\text{\textcolor{blue}{±1.94}}$} & 30.43\textsubscript{$_\text{\textcolor{blue}{±1.69}}$} & 16.39\textsubscript{$_\text{\textcolor{blue}{±0.33}}$} \\ \cmidrule(l){2-7} 
& LLaVA-1.5-13B$^\dagger$ & 5.00\textsubscript{$_\text{\textcolor{blue}{±0.39}}$} & 32.86\textsubscript{$_\text{\textcolor{blue}{±0.65}}$} & 120.49\textsubscript{$_\text{\textcolor{blue}{±2.28}}$} & 36.38\textsubscript{$_\text{\textcolor{blue}{±1.91}}$} & 10.11\textsubscript{$_\text{\textcolor{blue}{±0.36}}$} \\
& LLaVA-1.5-13B$^\star$ & 5.45\textsubscript{$_\text{\textcolor{blue}{±0.40}}$} & 39.91\textsubscript{$_\text{\textcolor{blue}{±0.64}}$} & 86.20\textsubscript{$_\text{\textcolor{blue}{±2.03}}$} & 71.16\textsubscript{$_\text{\textcolor{blue}{±1.85}}$} & 11.67\textsubscript{$_\text{\textcolor{blue}{±0.34}}$} \\ \midrule
\multirow{4}{*}{OmniDrive \cite{wang2025omnidrive}} 
& LLaMA-3-8B$^\dagger$ & 10.72\textsubscript{$_\text{\textcolor{blue}{±0.51}}$} & 43.45\textsubscript{$_\text{\textcolor{blue}{±0.56}}$} & \textbf{166.33}\textsubscript{$_\text{\textcolor{blue}{±2.47}}$} & 41.39\textsubscript{$_\text{\textcolor{blue}{±2.23}}$} & \textbf{19.38}\textsubscript{$_\text{\textcolor{blue}{±0.33}}$} \\
& LLaMA-3-8B$^\star$ & \textbf{12.99}\textsubscript{$_\text{\textcolor{blue}{±0.47}}$} & 42.43\textsubscript{$_\text{\textcolor{blue}{±0.79}}$} & 140.57\textsubscript{$_\text{\textcolor{blue}{±2.21}}$} & 51.49\textsubscript{$_\text{\textcolor{blue}{±2.14}}$} & 18.42\textsubscript{$_\text{\textcolor{blue}{±0.40}}$} \\ \cmidrule(l){2-7} 
& LLaVA-1.5-13B$^\dagger$ & 7.87\textsubscript{$_\text{\textcolor{blue}{±0.50}}$} & 40.12\textsubscript{$_\text{\textcolor{blue}{±0.84}}$} & 150.42\textsubscript{$_\text{\textcolor{blue}{±2.52}}$} & 47.56\textsubscript{$_\text{\textcolor{blue}{±2.34}}$} & 16.43\textsubscript{$_\text{\textcolor{blue}{±0.42}}$} \\
& LLaVA-1.5-13B$^\star$ & 5.14\textsubscript{$_\text{\textcolor{blue}{±0.48}}$} & \textbf{43.75}\textsubscript{$_\text{\textcolor{blue}{±0.80}}$} & 98.57\textsubscript{$_\text{\textcolor{blue}{±2.26}}$} & \textbf{86.10}\textsubscript{$_\text{\textcolor{blue}{±2.20}}$} & 13.24\textsubscript{$_\text{\textcolor{blue}{±0.40}}$} \\ \bottomrule[1.0pt]
\end{tabular}
}
\vspace{-0.12in}
\end{table}

\Tref{tab:main} presents the main results on \toolns, from which we derive the following insights:

\textbf{\ding{182} Comparison across \advlmns.}
The overall performance of all \advlm remains relatively low, indicating that current models still face notable limitations. OmniDrive consistently achieves the highest scores across all metrics, with a peak \textit{Success Rate} of 12.99\%, \textit{Driving Score} of 43.75, and \textit{Skill Score} of 19.38, indicating better robustness and generalization in closed-loop setting. In contrast, DriveLM and Dolphins show weaker performance, with average \textit{Driving Score} around 39.39 and \textit{Skill Score} below 18, reflecting limited planning reliability. EM-VLM4AD performs competitively, reaching a \textit{Comfortness} of 71.16, suggesting its lightweight design retains task-specific competence. Across all models, the standard deviations remain relatively small, typically under 5\% of each corresponding metric, indicating consistent and reliable performance across repeated trials.

\textbf{\ding{183} Comparison across parsing models.}
LLaMA and LLaVA are emphasized differently, with LLAMA having a relatively higher Success Rate and LLAVA having a relatively higher Driving Score. For example, under DriveLM with DCS, LLaMA achieves a \textit{Success Rate} of 10.45 and a \textit{Driving Score} of 38.79, while LLaVA yields only 5.00 and 42.06. 

\textbf{\ding{184} Comparison across parsing modes.}
Models using CNG generally achieve higher \textit{Driving Score} (43.45 \vs 42.43), as demonstrated by OmniDrive with LLaMA, reflecting the advantage of fine-grained control. On the other hand, DCS substantially improves \textit{Comfortness}, particularly for DriveLM with LLaMA (31.95 \vs 65.20) and Dolphins with LLaMA (37.85 \vs 43.54), suggesting better suppression of unstable behaviors across extended trajectories. (CNG \vs DCS). \textit{Detailed analysis for different modes can be found in the Supplementary Materials.}

\textbf{\ding{185} Micro-behavior Analysis.}
Under the same system configuration (\eg, LLaMA-3-8B with DCS), Dolphins achieves higher \textit{Efficiency} (132.58 \vs 107.44) compared to DriveLM, despite having a slightly lower \textit{Driving Score} (34.32 \vs 38.79). This suggests that while DriveLM is more focused on completing the route, Dolphins prioritizes smoother and more stable control. These results reflect differing behavioral tendencies: DriveLM emphasizes terminal task success, whereas Dolphins demonstrates finer control. \textit{Visualizations and failure analysis are in the supplementary materials.}

\begin{tcolorbox}
\vspace{-0.05in}
\textbf{Insight 1:} \advlm lack fine-grained control and show limited closed-loop performance, with low \textit{Success Rate} and \textit{Driving Score} highlighting a gap from deployment readiness.
\vspace{-0.05in}
\end{tcolorbox}

\subsection{Threat Scenario Evaluation}
\label{sec:Threat Scenario Evaluation}

\begin{table}[t]
\centering
\caption{Experimental results on \tool under threat scenarios. Blue subscripts denote the standard deviation (\textcolor{blue}{±std}) over multiple runs, while red superscripts indicate the performance drop (\textcolor{red}{-drop}) compared to the main results in \Tref{tab:main}.}
\label{tab:threat}
\resizebox{\textwidth}{!}{%
\begin{tabular}{@{}c|c|cc|cc|c@{}}
\toprule[1.0pt]
\multirow{2}{*}{\textbf{Fast System}} & \multirow{2}{*}{\textbf{Slow System}} & \multicolumn{2}{c|}{\textbf{Basic Performance}} & \multicolumn{2}{c|}{\textbf{Behavior Quality}} & \textbf{Specialized Capability} \\ \cmidrule(l){3-7} 
 &  & \textit{Success Rate} $\textcolor{blue}\uparrow$ & \textit{Driving Score} $\textcolor{blue}\uparrow$ & \textit{Efficiency} $\textcolor{blue}\uparrow$ & \textit{Comfortness} $\textcolor{blue}\uparrow$ & \textit{Skill Score} $\textcolor{blue}\uparrow$ \\ \midrule
\multirow{4}{*}{DriveLM \cite{drivelm}} 
& LLaMA-3-8B$^\dagger$ & 7.91\textsubscript{$_\text{\textcolor{blue}{±0.62}}^\text{\textcolor{red}{-1.18}}$} & 30.74\textsubscript{$_\text{\textcolor{blue}{±0.85}}^\text{\textcolor{red}{-8.67}}$} & 89.29\textsubscript{$_\text{\textcolor{blue}{±2.14}}^\text{\textcolor{red}{-38.27}}$} & 29.39\textsubscript{$_\text{\textcolor{blue}{±1.92}}^\text{\textcolor{red}{-2.56}}$} & 14.74\textsubscript{$_\text{\textcolor{blue}{±0.52}}^\text{\textcolor{red}{-2.81}}$} \\
& LLaMA-3-8B$^\star$ & 8.46\textsubscript{$_\text{\textcolor{blue}{±0.59}}^\text{\textcolor{red}{-1.99}}$} & 34.14\textsubscript{$_\text{\textcolor{blue}{±0.81}}^\text{\textcolor{red}{-4.65}}$} & 79.51\textsubscript{$_\text{\textcolor{blue}{±2.04}}^\text{\textcolor{red}{-27.93}}$} & 56.72\textsubscript{$_\text{\textcolor{blue}{±1.84}}^\text{\textcolor{red}{-8.48}}$} & 15.44\textsubscript{$_\text{\textcolor{blue}{±0.51}}^\text{\textcolor{red}{-1.16}}$} \\ \cmidrule(l){2-7}
& LLaVA-1.5-13B$^\dagger$ & 6.00\textsubscript{$_\text{\textcolor{blue}{±0.60}}^\text{\textcolor{red}{-0.82}}$} & 35.43\textsubscript{$_\text{\textcolor{blue}{±0.79}}^\text{\textcolor{red}{-1.86}}$} & 118.15\textsubscript{$_\text{\textcolor{blue}{±2.23}}^\text{\textcolor{red}{-11.68}}$} & 23.45\textsubscript{$_\text{\textcolor{blue}{±1.99}}^\text{\textcolor{red}{-5.50}}$} & 9.88\textsubscript{$_\text{\textcolor{blue}{±0.53}}^\text{\textcolor{red}{-4.23}}$} \\
& LLaVA-1.5-13B$^\star$ & 4.35\textsubscript{$_\text{\textcolor{blue}{±0.58}}^\text{\textcolor{red}{-0.65}}$} & 34.49\textsubscript{$_\text{\textcolor{blue}{±0.76}}^\text{\textcolor{red}{-7.57}}$} & 67.73\textsubscript{$_\text{\textcolor{blue}{±2.16}}^\text{\textcolor{red}{-18.01}}$} & 52.40\textsubscript{$_\text{\textcolor{blue}{±1.95}}^\text{\textcolor{red}{-17.47}}$} & 10.61\textsubscript{$_\text{\textcolor{blue}{±0.52}}^\text{\textcolor{red}{-2.49}}$} \\

\midrule
\multirow{4}{*}{Dolphins \cite{dolphins}} 
& LLaMA-3-8B$^\dagger$ & 7.91\textsubscript{$_\text{\textcolor{blue}{±0.67}}^\text{\textcolor{red}{-1.18}}$} & 30.53\textsubscript{$_\text{\textcolor{blue}{±0.90}}^\text{\textcolor{red}{-8.11}}$} & 110.63\textsubscript{$_\text{\textcolor{blue}{±2.19}}^\text{\textcolor{red}{-15.09}}$} & 27.63\textsubscript{$_\text{\textcolor{blue}{±1.97}}^\text{\textcolor{red}{-10.22}}$} & 14.63\textsubscript{$_\text{\textcolor{blue}{±0.57}}^\text{\textcolor{red}{-1.10}}$} \\
& LLaMA-3-8B$^\star$ & 7.70\textsubscript{$_\text{\textcolor{blue}{±0.64}}^\text{\textcolor{red}{-2.30}}$} & 30.89\textsubscript{$_\text{\textcolor{blue}{±0.86}}^\text{\textcolor{red}{-3.43}}$} & 102.09\textsubscript{$_\text{\textcolor{blue}{±2.09}}^\text{\textcolor{red}{-30.49}}$} & 35.70\textsubscript{$_\text{\textcolor{blue}{±1.89}}^\text{\textcolor{red}{-7.84}}$} & 16.04\textsubscript{$_\text{\textcolor{blue}{±0.56}}^\text{\textcolor{red}{-1.21}}$} \\ \cmidrule(l){2-7}
& LLaVA-1.5-13B$^\dagger$ & 7.12\textsubscript{$_\text{\textcolor{blue}{±0.65}}^\text{\textcolor{red}{-1.06}}$} & 31.27\textsubscript{$_\text{\textcolor{blue}{±0.84}}^\text{\textcolor{red}{-2.35}}$} & 111.01\textsubscript{$_\text{\textcolor{blue}{±2.28}}^\text{\textcolor{red}{-24.37}}$} & 47.64\textsubscript{$_\text{\textcolor{blue}{±2.04}}^\text{\textcolor{red}{-3.04}}$} & 12.18\textsubscript{$_\text{\textcolor{blue}{±0.58}}^\text{\textcolor{red}{-1.51}}$} \\
& LLaVA-1.5-13B$^\star$ & 1.99\textsubscript{$_\text{\textcolor{blue}{±0.63}}^\text{\textcolor{red}{-0.74}}$} & 32.13\textsubscript{$_\text{\textcolor{blue}{±0.81}}^\text{\textcolor{red}{-2.05}}$} & 54.35\textsubscript{$_\text{\textcolor{blue}{±2.21}}^\text{\textcolor{red}{-18.12}}$} & 58.63\textsubscript{$_\text{\textcolor{blue}{±2.00}}^\text{\textcolor{red}{-10.35}}$} & 5.01\textsubscript{$_\text{\textcolor{blue}{±0.57}}^\text{\textcolor{red}{-1.25}}$} \\

\midrule
\multirow{4}{*}{EM-VLM4AD \cite{gopalkrishnanmulti}} 
& LLaMA-3-8B$^\dagger$ & 7.61\textsubscript{$_\text{\textcolor{blue}{±0.42}}^\text{\textcolor{red}{-0.57}}$} & 32.66\textsubscript{$_\text{\textcolor{blue}{±0.59}}^\text{\textcolor{red}{-3.63}}$} & 122.87\textsubscript{$_\text{\textcolor{blue}{±1.89}}^\text{\textcolor{red}{-10.68}}$} & 27.03\textsubscript{$_\text{\textcolor{blue}{±1.67}}^\text{\textcolor{red}{-1.73}}$} & 16.12\textsubscript{$_\text{\textcolor{blue}{±0.32}}^\text{\textcolor{red}{-1.59}}$} \\
& LLaMA-3-8B$^\star$ & 9.50\textsubscript{$_\text{\textcolor{blue}{±0.39}}^\text{\textcolor{red}{-0.50}}$} & 33.70\textsubscript{$_\text{\textcolor{blue}{±0.61}}^\text{\textcolor{red}{-2.93}}$} & 111.18\textsubscript{$_\text{\textcolor{blue}{±1.79}}^\text{\textcolor{red}{-12.35}}$} & 27.39\textsubscript{$_\text{\textcolor{blue}{±1.59}}^\text{\textcolor{red}{-3.04}}$} & 15.57\textsubscript{$_\text{\textcolor{blue}{±0.31}}^\text{\textcolor{red}{-0.82}}$} \\ \cmidrule(l){2-7}
& LLaVA-1.5-13B$^\dagger$ & 4.55\textsubscript{$_\text{\textcolor{blue}{±0.40}}^\text{\textcolor{red}{-0.45}}$} & 31.22\textsubscript{$_\text{\textcolor{blue}{±0.59}}^\text{\textcolor{red}{-1.64}}$} & 110.85\textsubscript{$_\text{\textcolor{blue}{±1.98}}^\text{\textcolor{red}{-9.64}}$} & 33.83\textsubscript{$_\text{\textcolor{blue}{±1.74}}^\text{\textcolor{red}{-2.55}}$} & 9.60\textsubscript{$_\text{\textcolor{blue}{±0.33}}^\text{\textcolor{red}{-0.51}}$} \\
& LLaVA-1.5-13B$^\star$ & 4.96\textsubscript{$_\text{\textcolor{blue}{±0.38}}^\text{\textcolor{red}{-0.49}}$} & \textbf{37.12}\textsubscript{$_\text{\textcolor{blue}{±0.56}}^\text{\textcolor{red}{-2.79}}$} & 78.44\textsubscript{$_\text{\textcolor{blue}{±1.91}}^\text{\textcolor{red}{-7.76}}$} & \textbf{64.04}\textsubscript{$_\text{\textcolor{blue}{±1.70}}^\text{\textcolor{red}{-7.12}}$} & 10.62\textsubscript{$_\text{\textcolor{blue}{±0.32}}^\text{\textcolor{red}{-1.05}}$} \\

\midrule
\multirow{4}{*}{OmniDrive \cite{wang2025omnidrive}}
& LLaMA-3-8B$^\dagger$ & \textbf{10.18}\textsubscript{$_\text{\textcolor{blue}{±0.57}}^\text{\textcolor{red}{-0.54}}$} & 32.59\textsubscript{$_\text{\textcolor{blue}{±0.80}}^\text{\textcolor{red}{-10.86}}$} & 121.42\textsubscript{$_\text{\textcolor{blue}{±2.04}}^\text{\textcolor{red}{-44.91}}$} & 36.42\textsubscript{$_\text{\textcolor{blue}{±1.82}}^\text{\textcolor{red}{-4.97}}$} & \textbf{18.41}\textsubscript{$_\text{\textcolor{blue}{±0.47}}^\text{\textcolor{red}{-0.97}}$} \\
& LLaMA-3-8B$^\star$ & 9.87\textsubscript{$_\text{\textcolor{blue}{±0.54}}^\text{\textcolor{red}{-3.12}}$} & 31.40\textsubscript{$_\text{\textcolor{blue}{±0.76}}^\text{\textcolor{red}{-11.03}}$} & 116.68\textsubscript{$_\text{\textcolor{blue}{±1.94}}^\text{\textcolor{red}{-23.90}}$} & 36.05\textsubscript{$_\text{\textcolor{blue}{±1.74}}^\text{\textcolor{red}{-15.45}}$} & 16.39\textsubscript{$_\text{\textcolor{blue}{±0.46}}^\text{\textcolor{red}{-2.03}}$} \\ \cmidrule(l){2-7}
& LLaVA-1.5-13B$^\dagger$ & 7.32\textsubscript{$_\text{\textcolor{blue}{±0.55}}^\text{\textcolor{red}{-0.55}}$} & 36.51\textsubscript{$_\text{\textcolor{blue}{±0.74}}^\text{\textcolor{red}{-3.61}}$} & \textbf{124.85}\textsubscript{$_\text{\textcolor{blue}{±2.13}}^\text{\textcolor{red}{-25.57}}$} & 45.19\textsubscript{$_\text{\textcolor{blue}{±1.89}}^\text{\textcolor{red}{-2.38}}$} & 11.50\textsubscript{$_\text{\textcolor{blue}{±0.48}}^\text{\textcolor{red}{-4.93}}$} \\
& LLaVA-1.5-13B$^\star$ & 3.60\textsubscript{$_\text{\textcolor{blue}{±0.53}}^\text{\textcolor{red}{-1.54}}$} & 33.25\textsubscript{$_\text{\textcolor{blue}{±0.71}}^\text{\textcolor{red}{-10.50}}$} & 84.77\textsubscript{$_\text{\textcolor{blue}{±2.06}}^\text{\textcolor{red}{-13.80}}$} & 60.27\textsubscript{$_\text{\textcolor{blue}{±1.85}}^\text{\textcolor{red}{-25.83}}$} & 11.78\textsubscript{$_\text{\textcolor{blue}{±0.47}}^\text{\textcolor{red}{-1.46}}$} \\

\bottomrule[1.0pt]
\end{tabular}
\vspace{-0.1in}
}
\end{table}

\begin{figure}[!t]
    \centering
    \begin{subfigure}{0.24\textwidth}
        \centering
        \includegraphics[width=\textwidth]{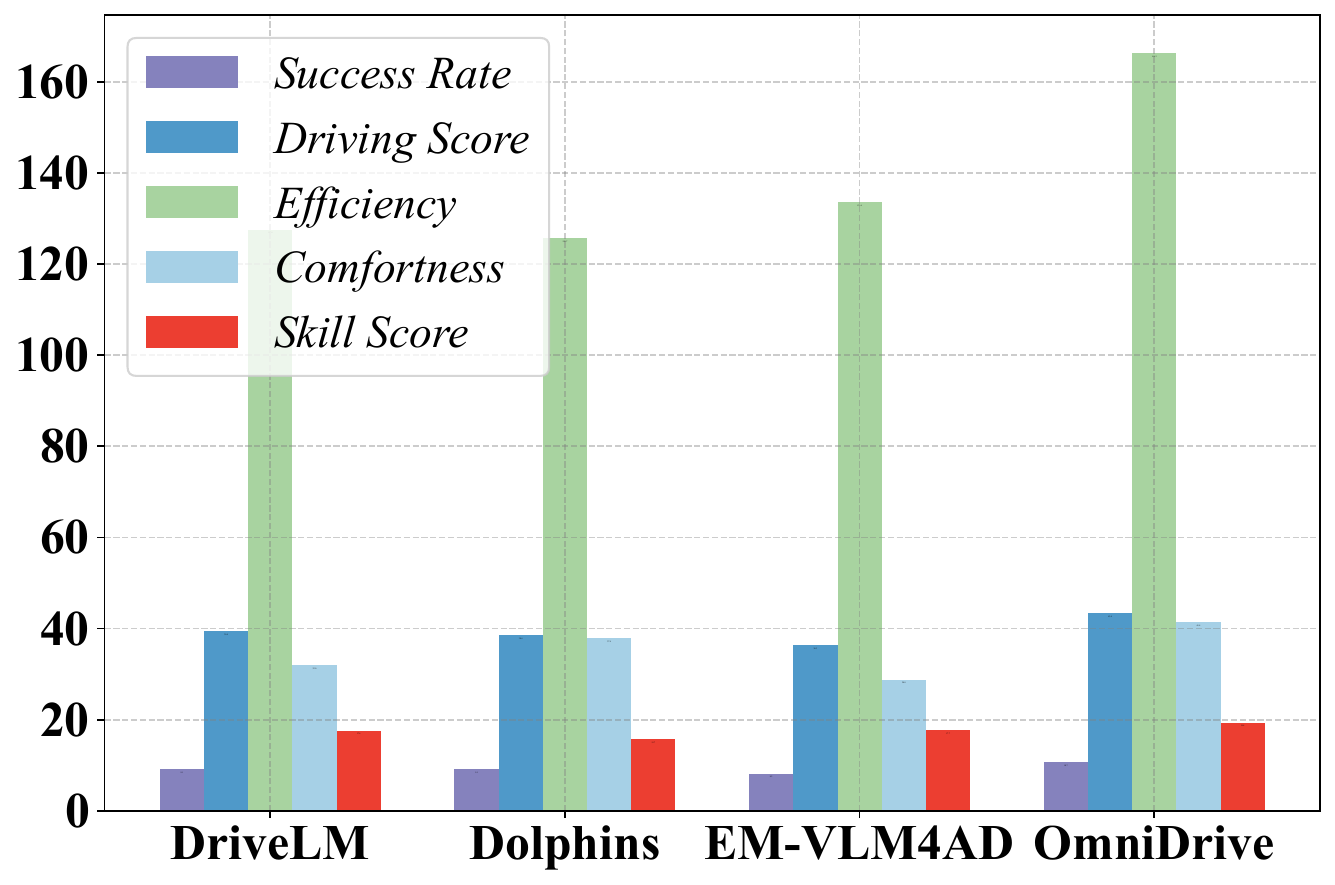}
        \caption{LLaMA + CNG setting in common scenarios.}
        \label{fig:main_llama}
    \end{subfigure}
    \hfill
    \begin{subfigure}{0.24\textwidth}
        \centering
        \includegraphics[width=\textwidth]{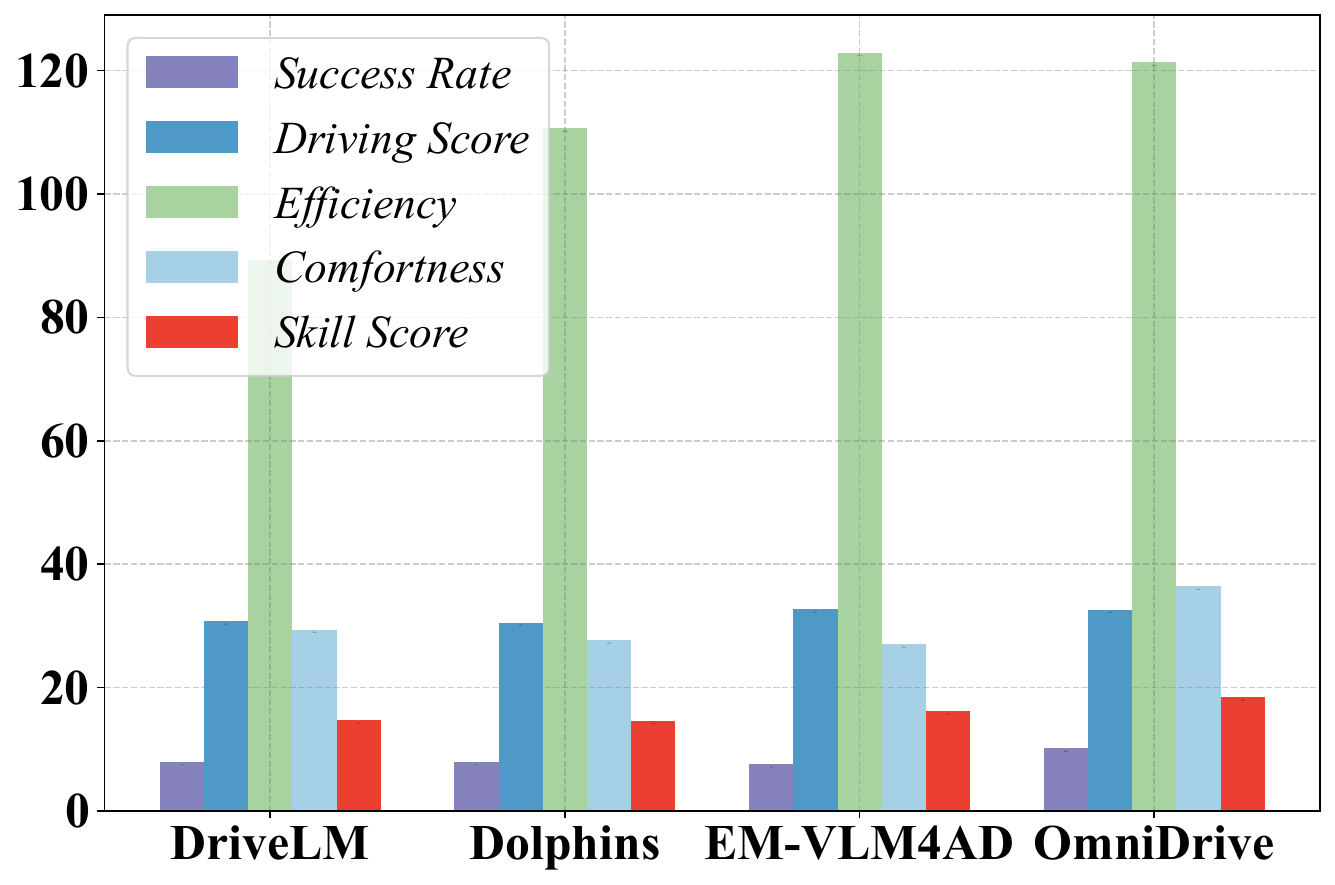}
        \caption{LLaMA + CNG setting in threat scenarios.}
        \label{fig:threat_llama}
    \end{subfigure}
    \hfill
    \begin{subfigure}{0.24\textwidth}
        \centering
        \includegraphics[width=\textwidth]{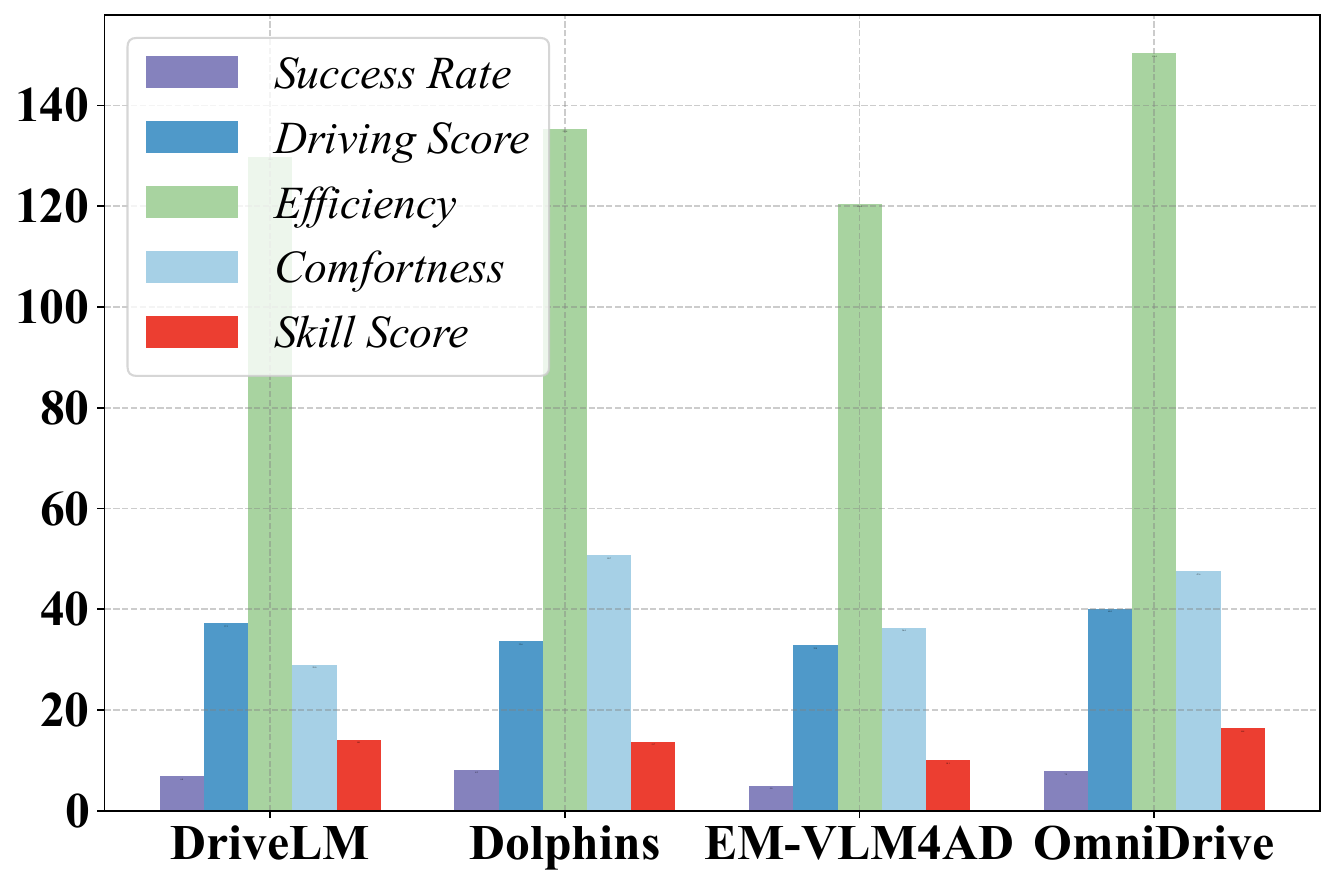}
        \caption{LLaVA + CNG setting in common scenarios.}
        \label{fig:main_llava}
    \end{subfigure}
    \hfill
    \begin{subfigure}{0.24\textwidth}
        \centering
        \includegraphics[width=\textwidth]{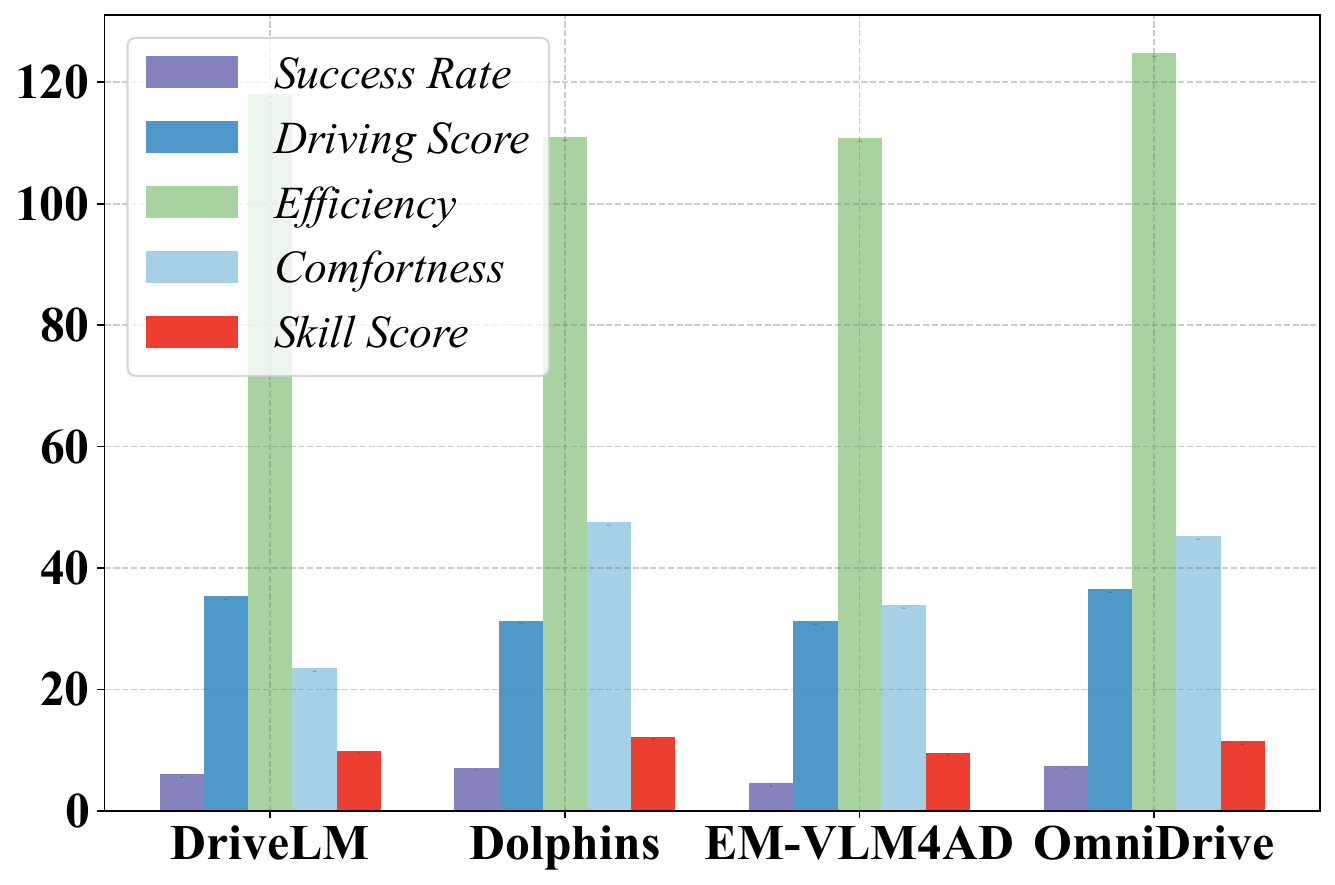}
        \caption{LLaVA + CNG setting in threat scenarios.}
        \label{fig:threat_llava}
    \end{subfigure}
    \caption{Model performance over different scenarios on \toolns.}
    \label{fig:bar}
    \vspace{-0.16in}
\end{figure}

\Tref{tab:threat} reports the results under threat scenarios, with a part of comparative visualization against the main experiment shown in \Fref{fig:bar}.

\textbf{\ding{182} Robustness degradation under threat scenarios.}
The overall performance trends remain consistent with the main results, but substantial degradations are observed across all models. For instance, the \textit{Success Rate} of Dolphins drops from 9.09\% to 7.91\% (1.18\%$\downarrow$), and EM-VLM4AD declines from 8.18\% to 7.61\% (0.57\%$\downarrow$) with the CNG mode and LLaMA. On average, the \textit{Driving Score} decreases by 26.1\%, and the \textit{Skill Score} drops by up to 3.59\%, especially for OmniDrive. 

\textbf{\ding{183} Stability Analysis via Standard Deviation.}
In addition to average performance degradation, we examine the variance across multiple runs to assess model stability under threat scenarios. OmniDrive, despite leading in all metrics under clean conditions, exhibits a noticeable increase in standard deviation across key indicators. Its \textit{Driving Score} variance rises from ±0.56 to ±0.80, and \textit{Skill Score} variance increases from ±0.33 to ±0.47. This suggests inconsistent behavior under adversarial perturbations.
In contrast, EM-VLM4AD maintains a relatively low and stable variance across all metrics. For instance, its \textit{Driving Score} standard deviation remains tightly bounded (±0.65 to ±0.59), and the \textit{Skill Score} variance stays around ±0.32, showing minimal fluctuation. The consistency underscores EM-VLM4AD's robustness and reliability in uncertain environments.

\textbf{\ding{184} Task-specific sensitivity under threat.}
Threat scenarios do not impact all task dimensions equally. 
Metrics related to smoothness and comfort (\eg, \textit{Efficiency} and \textit{Comfortness}) tend to degrade more sharply. 
For example, the \textit{Comfortness} of OmniDrive drops by 12.16 on average, while its \textit{Success Rate} decreases by just around 1.44\%. 
Similarly, DriveLM and Dolphins exhibit disproportionately larger declines in \textit{Efficiency} and \textit{Comfortness} than in \textit{Success Rate}, indicating that threats impact fine-grained behavior more than overall task success.

\begin{tcolorbox}
\vspace{-0.05in}
\textbf{Insight 2:} LLaVA shows a milder decline than LLaMA in behavior quality metrics (\eg, \textit{Efficiency}), while LLaMA performs better in basic performance (\eg, \textit{Success Rate}).
\vspace{-0.05in}
\end{tcolorbox}

\subsection{Real-world Evaluation}
\label{sec:Real-world Evaluation}

\begin{figure}[!t]
    \centering
    \begin{subfigure}{0.32\textwidth}
        \centering
        \includegraphics[width=\textwidth]{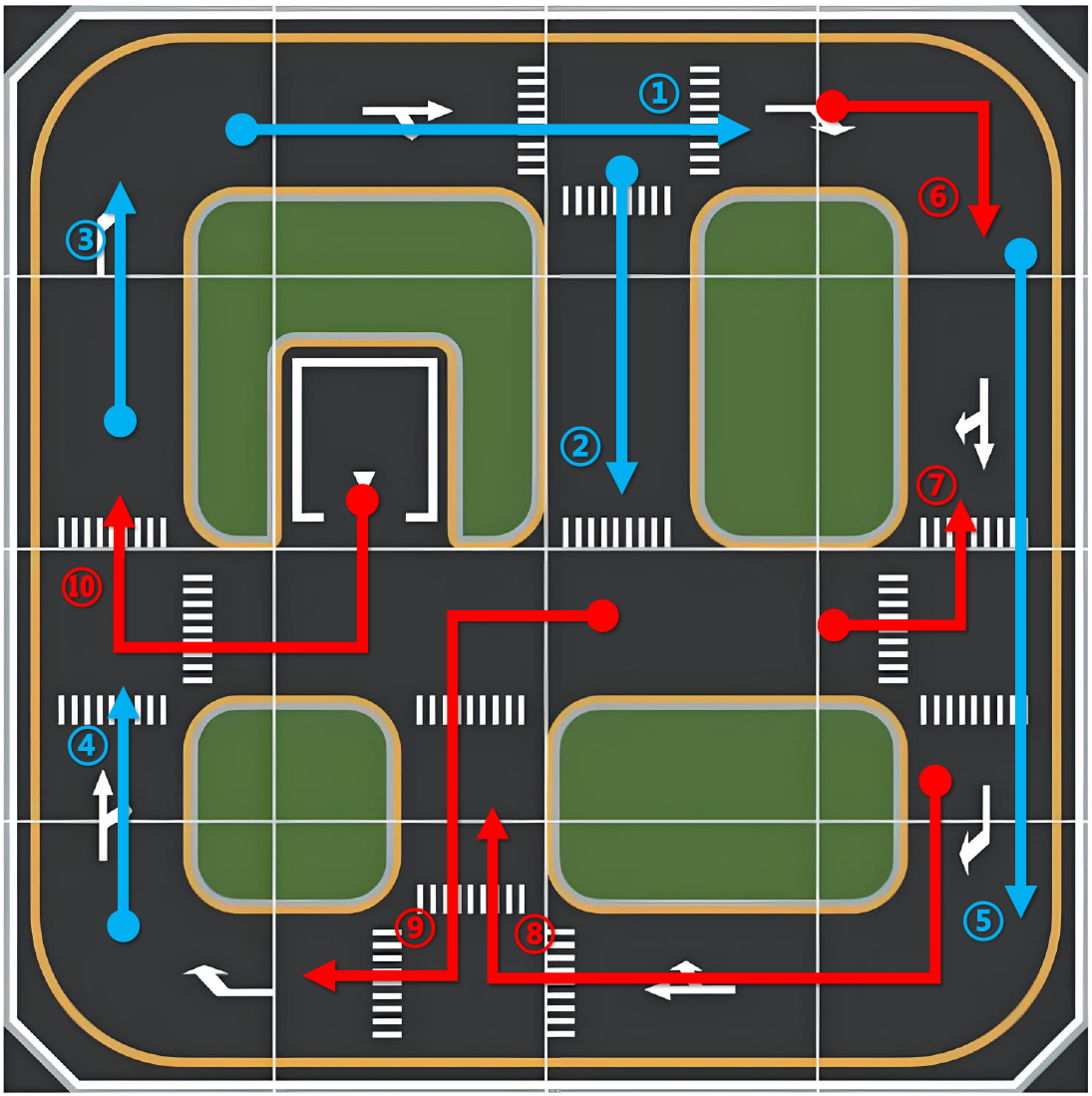}
        \caption{Illustration of the 10 routes: blue for straight, red for turns.}
        \label{fig:route}
    \end{subfigure}
    \hfill
    \begin{subfigure}{0.66\textwidth}
        \centering
        \includegraphics[width=0.95\linewidth]{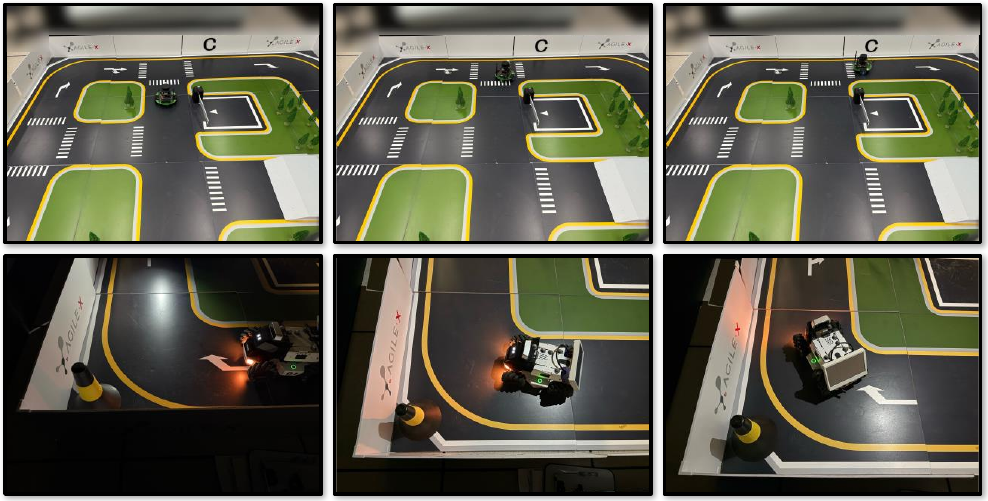}
        \caption{Real-world evaluation of \advlm using \toolns. Top: Jetbot (Route \textcircled{10}); Bottom: LIMO (Route \textcircled{6}).}
        \label{fig:real_world}
    \end{subfigure}
    \vspace{-0.03in}
    \caption{Real-world evaluation setup and representative experimental outcomes.}
    \label{fig:physical}

    \vspace{-0.10in}
\end{figure}

\textbf{Evaluation strategy.}
To quantitatively assess the real-world driving performance, we design a structured evaluation strategy centered on the lane-following task. 
The driving sandbox is partitioned into ten distinct route segments, each reflecting varying geometric and traffic complexities, as illustrated in \Fref{fig:route}. 
For each route, each AD vehicle runs three times, and we report the average results to ensure statistical reliability. 
The primary evaluation metric is the route completion rate, defined as the percentage of the planned trajectory successfully traversed by the vehicle without crossing the yellow boundary lines or colliding with obstacles. 

\begin{table}[!t]
\centering
\caption{Real-world experimental results on the Jetbot \cite{jetbot} and LIMO (Ackerman mode) \cite{limo} vehicle.}
\label{tab:real-exp}
\resizebox{\textwidth}{!}{%
\begin{tabular}{@{}c|cccccl|cccccl@{}}
\toprule[1.0pt]
\multirow{2}{*}{\advlmns} & \multicolumn{6}{c|}{\textbf{Jetbot}} & \multicolumn{6}{c}{\textbf{LIMO (Ackerman mode)}} \\ \cmidrule(l){2-13} 
 & route1 & route2 & route3 & route4 & route5 & \textit{Average} & route1 & route2 & route3 & route4 & route5 & \textit{Average} \\ \midrule
DriveLM \cite{drivelm} & 5/10 & 4/10 & 6/10 & 3/10 & 5/10 & 46\% & 4/10 & 5/10 & 3/10 & 4/10 & 5/10 & 42\% \\
Dolphins \cite{dolphins} & 4/10 & 3/10 & 5/10 & 2/10 & 3/10 & 34\% & 3/10 & 2/10 & 4/10 & 3/10 & 3/10 & 30\% \\
EM-VLM4AD \cite{gopalkrishnanmulti} & 6/10 & 6/10 & 7/10 & 5/10 & 6/10 & 60\% & 6/10 & 6/10 & 7/10 & 5/10 & 6/10 & 60\% \\
OmniDrive \cite{wang2025omnidrive} & 7/10 & 8/10 & 9/10 & 8/10 & 9/10 & 82\% & 8/10 & 9/10 & 8/10 & 9/10 & 9/10 & 86\% \\ \bottomrule[1.0pt]
\end{tabular}
}
\end{table}

\textbf{Results analysis.} 
As shown in \Tref{tab:real-exp}, all deployed models demonstrate the ability to complete basic lane-following tasks. 
In particular, the JetBot \cite{jetbot} platform consistently outperforms the LIMO \cite{limo} vehicle across all models, achieving an average route completion rate of 55.5\%, compared to 54.5\% on LIMO. 
This improvement partly stems from the simplicity of the evaluation routes.
However, even in these conditions, a significant number of failures still occur, highlighting persistent challenges in real-world deployment. 
\textit{Only the Ackerman mode of LIMO and 5 routes per vehicle are reported here. Additional modes, routes, and failure analyses are in the Supplementary Materials.}

\section{Potential Pathways}
Based on the above insights, we conduct preliminary experiments to explore the potential pathways for advancing robust and practical \advlm. \textit{Details are shown in Supplementary Materials}.

\textbf{Fine-Grained Control.} Insight 1 reveals that current \advlm lack fine-grained control, which shows significant limitations. Here, we aim to make the model output fine-grained control information for better execution in the closed-loop environment. Specifically, we enable the \advlm to simulate fine-grained control by selecting from predefined suffixes (\eg, extending ``Keep going straight'' with ``with speed = 0.5''). \textit{Selection rules are shown in the Supplementary Materials.} We use DriveLM with LLaMA, achieving a \textit{Success Rate} improvement from 9.09 to 12.52 on 220 standard routes in \toolns. This indicates that even simple fine-grained control can enhance performance. A more essential solution is to construct precise datasets incorporating control signals (\eg, historical and current control actions) and train the whole models, which we leave as future work.

\textbf{Hybrid Mode Switching.} Insight 2 suggests that different modes show different advantages over different tasks (LLaMA focuses on basic performance, while LLaVA emphasizes behavior quality). To leverage their strengths, we propose a hybrid mode switching strategy where \advlm identifies high-risk scenarios and selects LLaVA for risky cases and LLaMA otherwise. Specifically, we simply prompt the \advlm with ``Is there a security threat in the scene ahead?'' and select the mode based on its response (\ie, yes/no for LLaMA/LLaVA). Experiments on DriveLM on 220 standard routes in \tool demonstrate its effectiveness, improving both \textit{Success Rate} and \textit{Efficiency} (LLaMA: 9.09, 127.55; LLaVA: 6.82, 129.83; Hybrid: 9.23, 131.29). These results show that Hybrid Mode Switching effectively combines model strengths, enabling more adaptive and robust \advlmns.
\section{Conclusion and Future Work}
\label{sec:conclusion}

This paper introduces \toolns, a unified benchmark for closed-loop evaluation of \advlmns. \tool enables dynamic, real-time interaction by translating high-level driving commands into mid-level control actions via a dual-system adaptation architecture. A unified control abstraction layer further bridges these mid-level actions with low-level actuation on physical vehicles. \tool further incorporates self-reflective scenario generation targeting threat scenarios. Extensive experiments validate the effectiveness, providing insights for future research on \advlmns. \textbf{Limitations:} \ding{182} Real-world evaluation misses the complexity of commercial Autonomous Vehicles. \ding{183} Single-agent experiments lack interaction, limiting coordination assessment.

\textbf{Ethical Statement and Broader Impact.} This study uses no human subjects or sensitive data. All experiments are conducted in simulation or controlled settings. \tool is intended for research to improve AD performance, with no foreseeable ethical or societal concerns.

\medskip

\bibliographystyle{plain}
\bibliography{sample-base}





\end{document}